\documentclass[letterpaper, 10 pt, conference]{ieeeconf}  

\IEEEoverridecommandlockouts                              

\usepackage{amsfonts}
\usepackage[table]{xcolor}
\usepackage{graphicx}
\usepackage{booktabs}
\usepackage{comment}
\usepackage{censor}

\title{\LARGE \bf
Language-Conditioned Behavioral Cloning for Autonomous Endoscope Guidance
}

\author{Ariel Rodriguez$^{\dagger,1,4}$, Lorenzo Mazza$^{\dagger,1}$, \\Ortrun Hellig$^{2,3}$, Alejandra Toscano$^{1}$, Rayan Younis$^{2,3}$, Stefanie Speidel$^{1,2,4}$
\thanks{$\dagger$ These authors contributed equally to this work.}
\thanks{$^{1}$
NCT/UCC Dresden, UKDD Dresden, TU Dresden, DKFZ Heidelberg.}%
\thanks{$^{2}$The Center for Tactile Internet with Human-in-the-Loop (CeTI), TUD, Dresden, Germany.}%
\thanks{$^{3}$Department of Visceral, Thoracic and Vascular Surgery, Faculty of Medicine and University Hospital Carl Gustav Carus, TUD, Dresden, Germany.}%
\thanks{$^{4}$BMFTR Research Hub 6G-Life.}
}

\begin{document}

\maketitle
\thispagestyle{empty}
\pagestyle{empty}

\begin{abstract} 
Automated endoscope guidance reduces the need for continuous manual camera control during minimally invasive surgery. However, the desired camera motion depends on the surgeon’s immediate intentions and cannot always be inferred from the visual operative scene alone. We propose to automate the task using language-conditioned behavioral cloning; we define 6 clinically relevant guidance behaviors and collect 235 subtask demonstrations. In preliminary phantom experiments, our policy achieves 100\% success on focus, overview, and stop, 80\% on search-tip and follow-tip behaviors, and 50\% on horizon leveling. These results demonstrate the feasibility of learning heterogeneous surgeon-requested camera behaviors with a single visuomotor policy, while the observed failure modes motivate future work on adding temporal context for more demanding guidance tasks.
\end{abstract}

\section{INTRODUCTION}
In minimally invasive surgery, camera navigation determines the surgeon's view of the operative field and can directly affect procedural performance~\cite{huettl2020camera,ghaderi2021navigation}. Camera control also adds extra workload, either for an assisting operator in conventional laparoscopy or for the surgeon in robotic surgery~\cite{liu2023physical,weede2011}. Automated endoscope guidance aims to reduce the need for continuous manual camera control
while retaining the surgeon's ability to specify the desired view with minimal disruption to the surgical workflow. 
However, adapting in real time to different surgeon-requested camera behaviors and translating these requests into appropriate endoscope motion remains challenging.

Existing work has either relied on standard instrument tracking and visual servoing~\cite{omote1999,wijsman2018first}, or on language-based interfaces for selecting hand-designed camera-control behaviors~\cite{elazzazi2022natural}. More recent approaches have learned camera-control strategies while relying on a separate controller for execution~\cite{zhou2026strategy}, learned camera motion directly from visual observations without explicit task specification~\cite{li20223d}, or learned instruction-conditioned policies for a restricted class of tracking behaviors~\cite{ng2025endovla}. However, whether a single visuomotor policy can learn directly from demonstrations to execute a heterogeneous set of surgeon-requested laparoscopic camera-guidance behaviors remains underexplored.

We propose to automate endoscope guidance using language-conditioned behavioral cloning (BC).  The resulting transformer policy maps high-dimensional endoscopic observations together with real-time surgeon instructions directly to camera motion, without explicit instrument tracking or hand-designed, task-specific camera-control rules. 
The contributions of our work are the following: i) we formulate surgeon-requested endoscope guidance as a language-conditioned visuomotor control problem and address it with a transformer policy trained through BC; ii) we provide a clinically relevant protocol that frames the task as six explicit camera-motion primitives expressible as language instructions, and iii) we demonstrate with a preliminary evaluation that the resulting policy can successfully execute diverse laparoscopic camera-guidance behaviors.
\section{METHODS}
\subsection{Task protocol}
We define six endoscope-guidance subtasks intended to capture common camera adjustments required during minimally invasive surgery. The task set was developed together with clinicians and guided by established criteria for laparoscopic camera navigation, including target centering, viewing distance, horizon orientation, and instrument visibility~\cite{huettl2020camera,ghaderi2021navigation}. We additionally considered behaviors previously studied in automated camera-guidance systems, including instrument tracking, trajectory following, and maintaining or stopping camera motion~\cite{omote1999,weede2011,elazzazi2022natural}. The resulting protocol comprises \textit{focus}, \textit{overview}, \textit{search tip}, \textit{follow tip}, \textit{level horizon}, and \textit{stop}. Together, these motion primitives cover changes in viewing distance, instrument localization and tracking, viewpoint orientation, and deliberate camera stabilization. Each subtask is issued as an explicit guidance request, allowing different camera behaviors to be selected within the same surgical scene. Table~\ref{tab:task_definition} summarizes the desired endoscope response for each subtask.

 \begin{table}[t]
  \centering
  \caption{Endoscope-guidance tasks and their intended camera behavior.}
  \label{tab:task_definition}
  \small
  \renewcommand{\arraystretch}{1.35}
  \setlength{\tabcolsep}{5pt}
  \arrayrulecolor{gray!40}

  \begin{tabular}{p{0.23\columnwidth} p{0.68\columnwidth}}
  \hline
  \textbf{Task} & \textbf{Description} \\
  \hline
  \textit{Focus} &
  The instrument operator defines a working region by grasping two tissue segments or pointing with two instruments, with the space between their tips indicating the target. The endoscope moves toward the region while keeping it centered in the image. \\
  \hline
  \textit{Overview} &
  The instrument operator continues the current surgical activity, while the endoscope moves away from the observed working area to provide a wider view and keeps that area centered in the image. \\
  \hline
  \textit{Search tip} &
  The instrument operator indicates a target with one instrument. The endoscope searches for the instrument when it is outside the field of view, follows its visible shaft until the tip is located, and centers the midpoint between the instrument tip and the indicated target. \\
  \hline
  \textit{Follow tip} &
  The instrument operator points toward an anatomical structure and moves an instrument along its contour. The endoscope follows the instrument tip, keeping it near the center of the image while maintaining visibility of the
  surrounding anatomy. \\
  \hline
  \textit{Level horizon} &
  The endoscope operator rolls the endoscope to correct the camera orientation and restore a level horizon and upright view of the operative field. \\
  \hline
  \textit{Stop} &
  The other operators may continue activity in the phantom, remain inactive, or work outside the field of view, while the endoscope remains stationary and preserves the current view. \\
  \hline
  \end{tabular}
  \end{table}
\subsection{Hardware and Experimental Setup}

\begin{figure}[t]
  \centering
  \includegraphics[width=0.4\columnwidth]{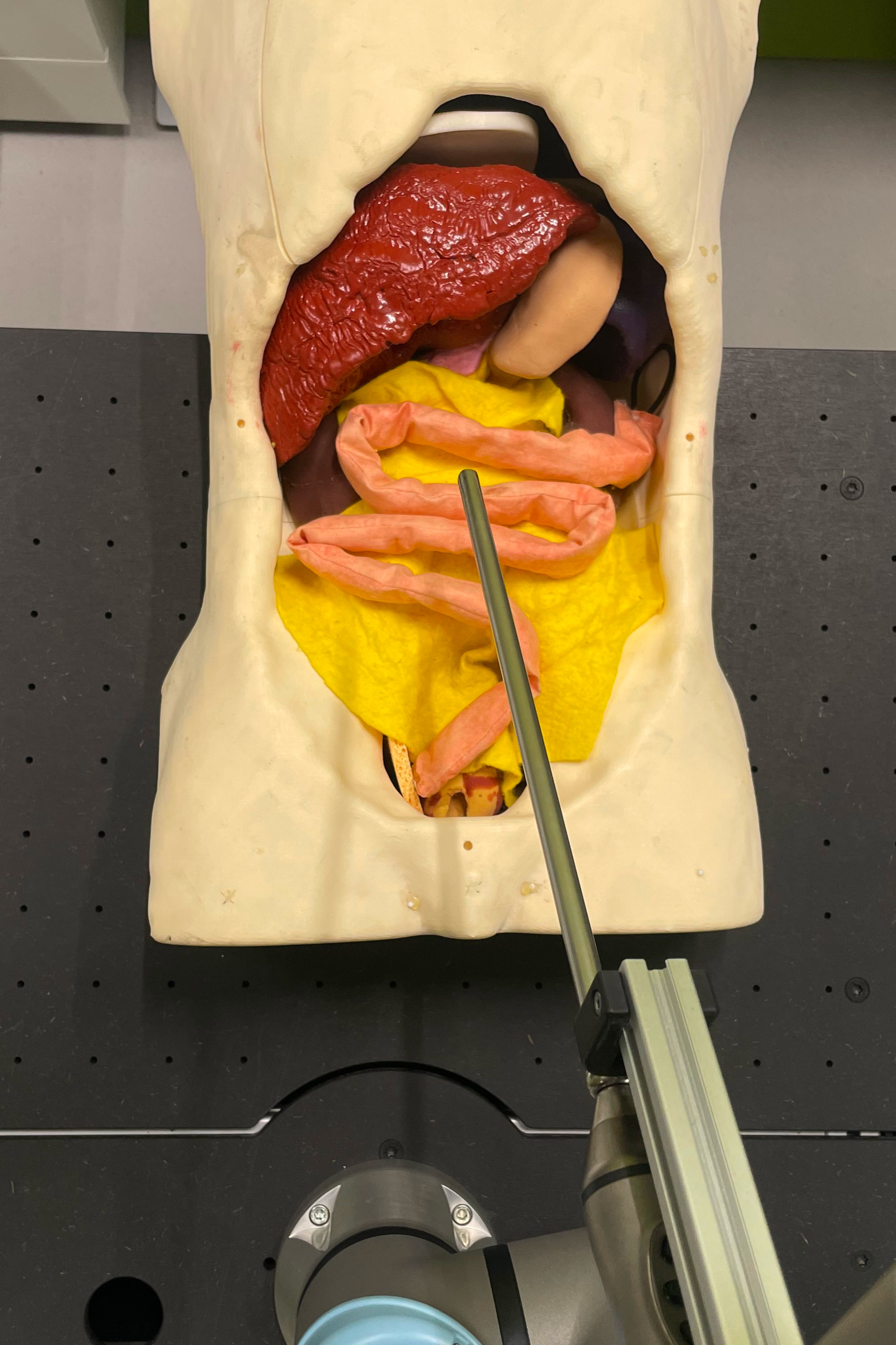}
  \caption{Environment for the endoscope-guidance task. The OpenHELP phantom is shown in the center, with the endoscope approaching from above. The setup was used for demonstration collection and policy deployment.}
  \label{fig:environment}
\end{figure}

We use the robotic platform for autonomous laparoscopic surgery described in \cite{rodriguez2026open_source}. The experimental setup consists of a UR5e robotic arm equipped with a stereo TIPCAM1 S 3D endoscope (Karl Storz SE \& Co. KG) and the OpenHELP phantom \cite{hanna2015openhelp}. The robot is controlled through a joystick-based interface while enforcing the remote center-of-motion (RCM) constraint. The experimental setup is shown in Figure~\ref{fig:environment}.

\subsection{Data Collection}
\label{sec:data_collection}

Data was collected with three operators: one controlled the robotic endoscope, another operated two laparoscopic instruments that provided visual references for selected tasks, and a third operated an electrocautery instrument in selected episodes to reproduce scenes containing three instruments. Each episode corresponded to one guidance task. We recorded synchronized stereo images, the three-dimensional endoscope-tip position, the endoscope roll angle, and one language instruction per subtask. The dataset contains 235 episodes, which result in a recording time of 39:53 (minutes:seconds),  distributed across \textit{focus} (6:20), ~\textit{overview} (6:01), ~\textit{level horizon} (7:11), ~\textit{follow tip} (4:55), ~\textit{search tip} (6:18), and \textit{stop} (9:08).

\subsection{Task Space and Policy Definition}

At each time step $t$, the policy receives the synchronized stereo endoscopic image pair and the language instruction associated with the current subtask, $o_t = (I_t^{\mathrm{left}}, I_t^{\mathrm{right}}, \ell_t)$.
The action $a_t = (\Delta p_t, \Delta r_t)$, consists of the relative 3D displacement of the endoscope tip, $\Delta p_t \in \mathbb{R}^{3}$, and the change in endoscope roll angle, $\Delta r_t \in \mathbb{R}$. Our policy is a language-conditioned behavioral-cloning decoder-only action transformer (BCAT) that maps the stereo endoscopic image pair and language instruction at time $t$ to a chunk of $H=10$ future actions $a_{t:t+H}$. Separate pretrained EfficientNet-B0 encoders process the left and right endoscopic images and are fine-tuned during training \cite{tan2019efficientnet}. The language instruction is encoded using a frozen pretrained MiniLM-L6 encoder \cite{wang2020minilm}. The bidirectional transformer decoder uses $H$ learnable action queries that cross-attend to the visual and language observation tokens and predict the action chunk in parallel. The policy is trained by minimizing the $\ell_1$ error between the predicted and demonstrated action chunks.
\begin{figure*}[!t]
  \centering
  \includegraphics[width=\textwidth]{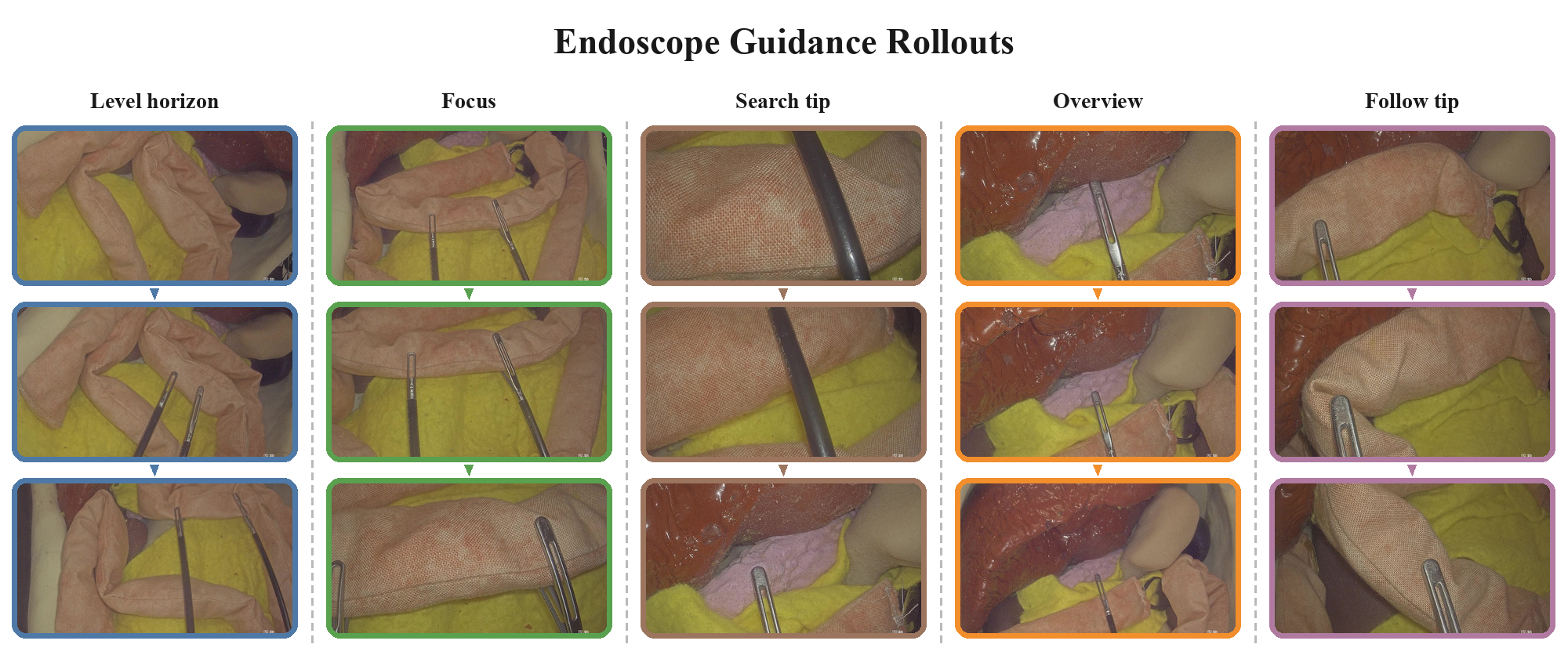}
  \caption{Representative policy rollouts for five tasks. Each column shows the temporal evolution of one task from top to bottom.}
  \label{fig:rollouts}
\end{figure*}

\begin{table*}[t]
    \centering
    \caption{Evaluation of the BC policy on the guidance tasks, with ten
    rollouts per task. Timing statistics are averaged over successful
    sub-executions and reference human expert demonstrations.
    The completion time ratio is the average human completion time divided
    by the average policy completion time. Timing metrics are not applicable
    to the Stop task.}
    \label{tab:success_rates}
    \small
    \renewcommand{\arraystretch}{1.2}
    \begin{tabular}{lcccccc}
      \toprule
      & \textbf{Focus} & \textbf{Overview} & \textbf{Search tip}
      & \textbf{Follow tip} & \textbf{Level horizon} & \textbf{Stop} \\
      \midrule
      Success rate (\%, $\uparrow$)
        & 100 (10/10) & 100 (10/10) & 80 (8/10)
        & 80 (8/10) & 50 (5/10) & 100 (10/10) \\
      Average human completion time (s)
        & 7.38 & 6.91 & 8.02 & 13.06 & 9.70 & -- \\
      Average policy completion time (s)
        & 9.36 & 7.40 & 15.34 & 34.71 & 35.95 & -- \\
      Completion time ratio ($\uparrow$)
        & 0.79 & 0.93 & 0.52 & 0.38 & 0.27 & -- \\
      \bottomrule
    \end{tabular}
  \end{table*}
\section{Experiments and Results}

We train the policy for 100 epochs and evaluate the final checkpoint.
We evaluate the final checkpoint over 10 rollouts with varied initial conditions, each comprising the 6 guidance subtasks, for a total of 60 subtask evaluations. The operator triggers each task through a spoken instruction, which is transcribed and provided to the policy. Each execution is scored against the corresponding task definition and considered a failure if it exceeds a maximum timestep. Table~\ref{tab:success_rates} reports the quantitative results, while Fig.~\ref{fig:rollouts} shows representative rollouts. The policy achieves 100\% success on \textit{focus}, \textit{overview}, and \textit{stop}, and 80\% on both \textit{search tip} and \textit{follow tip}. \textit{Level horizon} is the most challenging task, with a 50\% success rate. Execution speed is close to the demonstrations for \textit{overview}, while larger differences occur for tasks requiring longer-term search, tracking, or orientation correction.
The main failure modes occurred during \textit{level horizon}, \textit{search tip}, and \textit{follow tip}. Because the policy observes only the current frame, it cannot explicitly infer recent camera motion or retain information about previously observed regions. This can limit search and tracking subtasks and can lead to repeated corrections. \textit{Level horizon} is additionally affected by the lack of an unambiguous visual reference for the desired scene orientation.

\section{Conclusion}
We presented a language-conditioned behavioral-cloning approach for surgeon-requested endoscope guidance and evaluated it across six camera behaviors. Preliminary phantom experiments show that a single visuomotor policy can successfully execute several heterogeneous guidance tasks directly from endoscopic observations and language instructions. Performance remains limited for tasks requiring temporal context, particularly search, tracking, and horizon correction. Future work will incorporate observation history and evaluate more expressive policy parameterizations, including generative and larger vision-language-action models.

\clearpage

\section*{ACKNOWLEDGMENT}

Lorenzo Mazza, Alejandra Toscano and Stefanie Speidel acknowledge support by the project ”Next Generation AI Computing (gAIn),” funded by the Bavarian Ministry of Science and the Arts and the Saxon Ministry for Science, Culture, and Tourism as well as by the Hightech Agenda Bavaria. 

Ariel Rodriguez, Ortrun Hellig, Rayan Younis and Stefanie Speidel acknowledge the support by the German Research Foundation (DFG, Deutsche Forschungsgemeinschaft) as part of Germany’s Excellence Strategy – EXC 2050/1 – Project ID 390696704 – Cluster of Excellence “Centre for Tactile Internet with Human-in-the-Loop” (CeTI) of Technische Universität Dresden.

Ariel Rodriguez and Stefanie Speidel would additionally like to thank the Federal Ministry of Research, Technology, and Space (BMFTR) for its support as part of the research program Communication Systems “Souverän. Digital. Vernetzt”. Joint project 6G-life, project identification number: 16KIS2413K. 

This work has been partially supported by the German Federal Ministry of Research, Technology and Space (BMFTR) under the Robotics Institute Germany (RIG).

\bibliographystyle{IEEEtran}
\bibliography{references}

@article{huettl2020camera,
    author  = {Huettl, Florentine and Huber, Tobias and Duwe, Matthias and
               Lang, Hauke and Paschold, Markus and Kneist, Werner},
    title   = {Higher Quality Camera Navigation Improves the Surgeon's Performance:
               Evidence from a Pre-Clinical Study},
    journal = {Journal of Minimal Access Surgery},
    volume  = {16},
    number  = {4},
    pages   = {355--359},
    year    = {2020},
    doi     = {10.4103/jmas.JMAS_143_19}
}

@inproceedings{li20223d,
  title={3D perception based imitation learning under limited demonstration for laparoscope control in robotic surgery},
  author={Li, Bin and Wei, Ruofeng and Xu, Jiaqi and Lu, Bo and Yee, Chi Hang and Ng, Chi Fai and Heng, Pheng-Ann and Dou, Qi and Liu, Yun-Hui},
  booktitle={2022 International Conference on Robotics and Automation (ICRA)},
  pages={7664--7670},
  year={2022},
  organization={IEEE}
}

@inproceedings{ng2025endovla,
  title={Endovla: Dual-phase vision-language-action for precise autonomous tracking in endoscopy},
  author={Ng, Chi Kit and Bai, Long and Wang, Guankun and Wang, Yupeng and Gao, Huxin and Jin, Chenhan and Zeng, Tieyong and Ren, Hongliang and others},
  booktitle={9th Annual Conference on Robot Learning},
  year={2025}
}

@article{wijsman2018first,
  title={First experience with THE AUTOLAP™ SYSTEM: an image-based robotic camera steering device},
  author={Wijsman, Paul JM and Broeders, Ivo AMJ and Brenkman, Hylke J and Szold, Amir and Forgione, Antonello and Schreuder, Henk WR and Consten, Esther CJ and Draaisma, Werner A and Verheijen, Paul M and Ruurda, Jelle P and others},
  journal={Surgical endoscopy},
  volume={32},
  number={5},
  pages={2560--2566},
  year={2018},
  publisher={Springer}
}

@article{ghaderi2021navigation,
    author  = {Ghaderi, Iman and Hsu, Chiu-Hsieh and Hines, Eric M. and
               Alabagi, Abdulla and Galvani, Carlos C.},
    title   = {The Impact of Navigation Grid Overlay on Performance of Camera
               Assistants during Laparoscopic Abdominal Procedures:
               A Randomized Controlled Trial},
    journal = {Journal of Surgical Education},
    volume  = {78},
    number  = {3},
    pages   = {991--997},
    year    = {2021},
    doi     = {10.1016/j.jsurg.2020.09.015}
}

@article{liu2023physical,
    author  = {Liu, Junjie and Qiao, Xi and Xiao, Yi and Deng, Zhuofan and
               Cui, Ji and Wu, Mingdong and Zhang, Haolong and Ran, Kun and
               Luo, Hailong and Tang, Bo},
    title   = {Physical and Mental Health Impairments Experienced by Operating
               Surgeons and Camera-Holder Assistants during Laparoscopic Surgery:
               A Cross-Sectional Survey},
    journal = {Frontiers in Public Health},
    volume  = {11},
    pages   = {1264642},
    year    = {2023},
    doi     = {10.3389/fpubh.2023.1264642}
}

@inproceedings{weede2011,
    author    = {Weede, Oliver and M{\"o}nnich, Heiko and M{\"u}ller, Bernd and
                 W{\"o}rn, Holger},
    title     = {An Intelligent and Autonomous Endoscopic Guidance System for
                 Minimally Invasive Surgery},
    booktitle = {2011 IEEE International Conference on Robotics and Automation},
    pages     = {5762--5768},
    year      = {2011},
    doi       = {10.1109/ICRA.2011.5980216}
}

@article{omote1999,
    author  = {Omote, K. and Feussner, H. and Ungeheuer, A. and Arbter, K. and
               Wei, G. Q. and Siewert, J. R. and Hirzinger, G.},
    title   = {Self-Guided Robotic Camera Control for Laparoscopic Surgery
               Compared with Human Camera Control},
    journal = {The American Journal of Surgery},
    volume  = {177},
    number  = {4},
    pages   = {321--324},
    year    = {1999},
    doi     = {10.1016/S0002-9610(99)00055-0}
}

@article{zhou2026strategy,
    author  = {Zhou, Keyu and Xu, Peisen and Wu, Yahao and Chen, Jiming and
               Li, Gaofeng and Li, Shunlei},
    title   = {Strategy-Supervised Autonomous Laparoscopic Camera Control via
               Event-Driven Graph Mining},
    journal = {arXiv preprint arXiv:2602.20500},
    year    = {2026},
    eprint  = {2602.20500},
    archivePrefix = {arXiv}
}

@article{elazzazi2022natural,
    author  = {Elazzazi, Maysara and Jawad, Luay and Hilfi, Mohammed and
               Pandya, Abhilash},
    title   = {A Natural Language Interface for an Autonomous Camera Control
               System on the da Vinci Surgical Robot},
    journal = {Robotics},
    volume  = {11},
    number  = {2},
    pages   = {40},
    year    = {2022},
    doi     = {10.3390/robotics11020040}
}

@misc{rodriguez2026open_source,
    author       = {Rodriguez, Ariel and Mazza, Lorenzo and Lelis, Martin and
                    Younis, Rayan and Bodenstedt, Sebastian and Wagner, Martin
                    and Speidel, Stefanie},
    title        = {An Open-Source Robotics Research Platform for Autonomous
                    Laparoscopic Surgery},
    year         = {2026},
    eprint       = {2603.08490},
    archivePrefix = {arXiv},
    primaryClass = {cs.RO}
  }

@article{hanna2015openhelp,
    author  = {Kenngott, H. G. and W{\"u}nscher, J. J. and Wagner, M. and
               Preukschas, A. and Wekerle, A. L. and Neher, P. and Suwelack, S.
               and Speidel, S. and Nickel, F. and Oladokun, D. and Albala, L.
               and Maier-Hein, L. and Dillmann, R. and Meinzer, H. P. and
               M{\"u}ller-Stich, B. P.},
    title   = {OpenHELP (Heidelberg Laparoscopy Phantom): Development of an
               Open-Source Surgical Evaluation and Training Tool},
    journal = {Surgical Endoscopy},
    volume  = {29},
    number  = {11},
    pages   = {3338--3347},
    year    = {2015},
    doi     = {10.1007/s00464-015-4094-0}
  }

@inproceedings{tan2019efficientnet,
    author    = {Tan, Mingxing and Le, Quoc V.},
    title     = {EfficientNet: Rethinking Model Scaling for Convolutional Neural
                 Networks},
    booktitle = {Proceedings of the 36th International Conference on Machine
                 Learning},
    pages     = {6105--6114},
    year      = {2019}
  }

@inproceedings{wang2020minilm,
    author    = {Wang, Wenhui and Wei, Furu and Dong, Li and Bao, Hangbo and
                 Yang, Nan and Zhou, Ming},
    title     = {MiniLM: Deep Self-Attention Distillation for Task-Agnostic
                 Compression of Pre-Trained Transformers},
    booktitle = {Advances in Neural Information Processing Systems},
    volume    = {33},
    year      = {2020}
  }

\end{document}